\documentclass[letterpaper, 10 pt, conference]{ieeeconf} 

\IEEEoverridecommandlockouts                              
                                                          
\usepackage{graphicx} % for pdf, bitmapped graphics files
\usepackage{amsmath} % assumes amsmath package installed
\usepackage{amssymb}  % assumes amsmath package installed
\usepackage{url}
\usepackage{hyperref}
\usepackage{bm}
\usepackage[cmyk,table,hyperref]{xcolor}

\usepackage{balance}

\DeclareMathOperator{\Sim}{Sim}
\DeclareMathOperator{\SE}{SE}

\title{\LARGE \bf
LDSO: Direct Sparse Odometry with Loop Closure
}

\author{Xiang Gao, Rui Wang, Nikolaus Demmel and Daniel Cremers% <-this % stops a space
\thanks{Xiang Gao, Rui Wang, Nikolaus Demmel and Daniel Cremers are with the Computer Vision %
Group, Department of Informatics, Technical University of Munich, Germany %
        {\tt\small \{gaoxi,wangr,demmeln,cremers\}@in.tum.de}}%
\thanks{%
This work was partially supported by the grant ``For3D'' by the Bavarian Research Foundation, the grant CR~250/9-2 ``Mapping on Demand'' by the German Research Foundation and the ERC Consolidator Grant 649 323 ``3D Reloaded''.%
}}% According to Sabine there is no number for For3D

\begin{document}
\maketitle
\thispagestyle{empty}
\pagestyle{empty}

%%%%%%%%%%%%%%%%%%%%%%%%%%%%%%%%%%%%%%%%%%%%%%%%%%%%%%%%%%%%%%%%%%%%%%%%%%%%%%%%
\begin{abstract}
In this paper we present an extension of Direct Sparse Odometry (DSO) \cite{Engel2018} to a monocular visual SLAM system with loop closure detection and pose-graph optimization (LDSO).
As a direct technique, DSO can utilize any image pixel with sufficient intensity gradient, which makes it robust even in featureless areas.
LDSO retains this robustness, while at the same time ensuring repeatability of some of these points by favoring corner features in the tracking frontend. This repeatability allows to reliably detect loop closure candidates with a conventional feature-based bag-of-words (BoW) approach.
Loop closure candidates are verified geometrically and Sim(3) relative pose constraints are estimated by jointly minimizing 2D and 3D geometric error terms.
These constraints are fused with a co-visibility graph of relative poses extracted from DSO's sliding window optimization.
Our evaluation on publicly available datasets demonstrates that the modified point selection strategy retains the tracking accuracy and robustness, and the integrated pose-graph optimization significantly reduces the accumulated rotation-, translation- and scale-drift, resulting in an overall performance comparable to state-of-the-art feature-based systems, even without global bundle adjustment.
\end{abstract}

%%%%%%%%%%%%%%%%%%%%%%%%%%%%%%%%%%%%%%%%%%%%%%%%%%%%%%%%%%%%%%%%%%%%%%%%%%%%%%%%
\section{INTRODUCTION}
Simultaneous Localization and Mapping (SLAM) has been an active research area in computer vision and robotics for several decades since the 1980s \cite{Cadena2016,Fuentes-Pacheco2015}. 
It is a fundamental module of many applications that need real-time localization like mobile robotics, autonomous MAVs, autonomous driving, as well as virtual and augmented reality \cite{Klein2007}.
While other sensor modalities such as laser scanners, GPS, or inertial sensors are also commonly used, visual SLAM has been very popular, in part because cameras are readily available in consumer products and passively acquire rich information about the environment.
In particular, in this work we focus on the monocular case of tracking a single gray-scale camera.

Typically, a visual SLAM system consists of a camera tracking frontend, and a backend that creates and maintains a map of keyframes, and reduces global drift by loop closure detection and map optimization. 
The frontend may localize the camera globally against the current map \cite{Klein2007,Mur-Artal2015}, track the camera locally with visual (keyframe) odometry (VO) \cite{Kerl2013,Engel2014}, or use a combination of both \cite{Lynen2015,Qin2017,Schneider2018}.

There are several open challenges in adapting a direct, sliding-window, marginalizing odometry system like DSO to reuse existing information from a map.
For example, in order to evaluate the photometric error, images of past keyframes would have to be kept in memory, and when incorporating measurements from previous keyframes, it is challenging to ensure estimator consistency, since information from these keyframes that is already contained in the marginalization prior should not be reused.
We therefore propose to adapt DSO as our SLAM frontend to estimate visual odometry with local consistency and correct its drift with loop closure detection and pose graph optimization in the backend.
Note that DSO itself consists also of a camera-tracking frontend and a backend that optimizes keyframes and point depths. 
However, in this work we refer to the whole of DSO as our odometry frontend.

\begin{figure}[!t]
	\centering
	\includegraphics[width=0.48\textwidth]{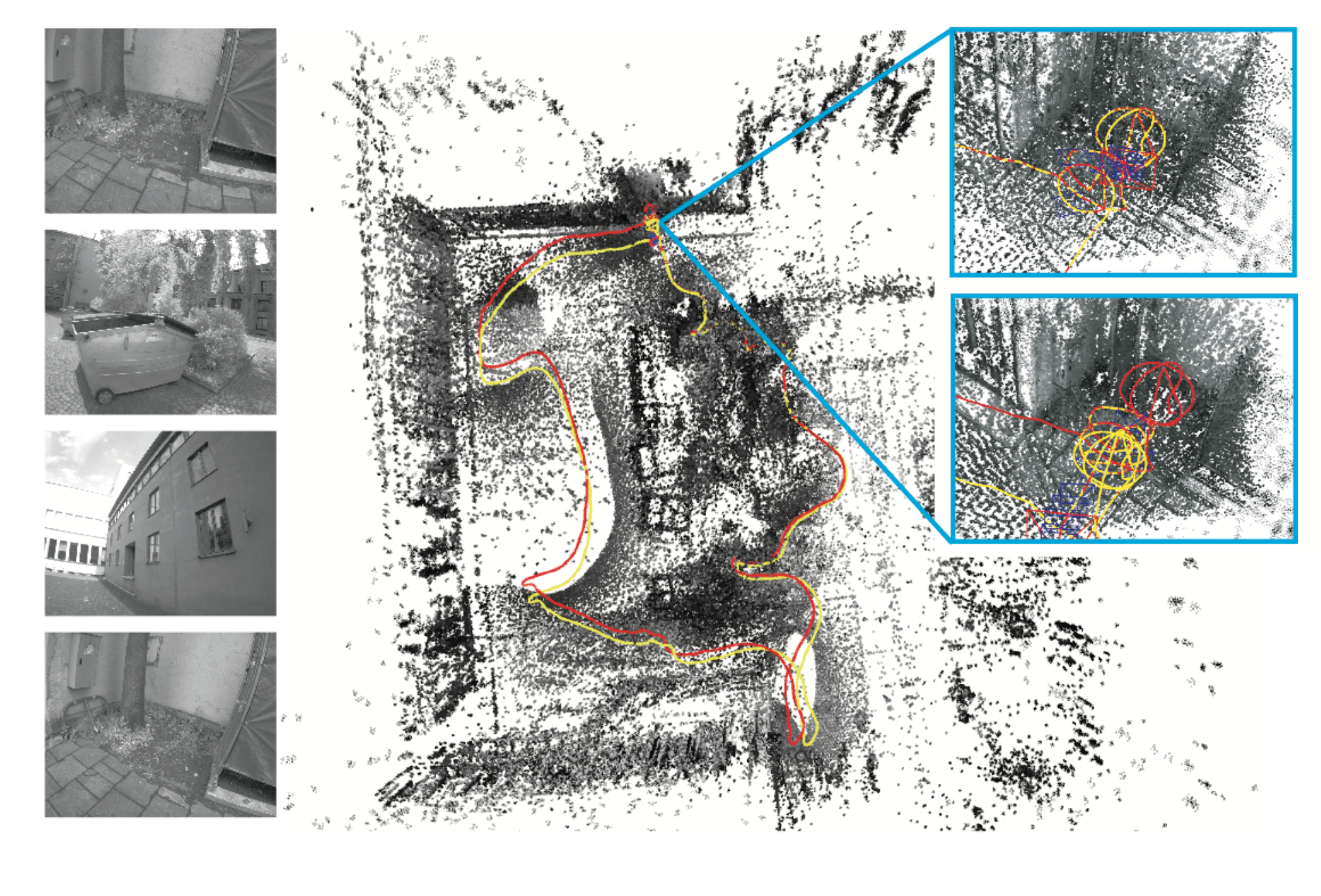}
	\vspace{-1em}
	\caption{Estimated trajectories with and without loop closing in the 
	TUM-Mono dataset. The left part are sample images from 
	sequence 31, whose end point should be at the same place as the 
	start point. The right part shows the estimated trajectory by LDSO before (red) and after (yellow) loop closure. 
	The zoom-in highlights how trajectories and point-clouds align much better after.
	}
	\label{fig:1}
	\vspace{-1em}
\end{figure}

VO approaches can be divided into two categories: indirect (feature-based) methods that minimize the reprojection error with fixed, previously estimated correspondences between repeatable discrete feature points, and direct methods that jointly estimate motion and correspondences by minimizing the photometric error in direct image alignment. 
While feature-based methods have been the mainstream for a long time, recent advances in direct VO have shown better accuracy and robustness, especially when the images do not contain enough explicit corner features \cite{Engel2018, Yang2017}. 
The robustness in the direct approach comes from the joint estimation of motion and correspondences as well as the ability to also use non-corner pixels, corresponding to edges, or even smooth image regions (as long as there is sufficient image gradient).
However, without loop closing, both indirect and direct VO suffers from the accumulated drift in the unobservable degrees-of-freedom, which are global translation, rotation and scale in the monocular case.
This makes the long-term camera trajectory and map inaccurate and thus limits the application to only short-term motion estimation. 

In order to close loops, they need to be detected first.
The state-of-the-art loop detection methods --- sometimes referred to as appearance-only SLAM --- are usually based on indexed image features (e.g. BoW \cite{Botterill2011,Filliat2007,Galvez-Lopez2012,Cummins2008}) and thus can be directly applied in feature-based VO by reusing the features from the frontend.
This, however, is not as straightforward in the direct case: 
If we detect and match features independently from the frontend, we might not have depth estimates for those points, which we need to efficiently estimate $\Sim(3)$ pose-constraints, and if instead we attempt to reuse the points from the frontend and compute descriptors for those, they likely do not correspond to repeatable features and lead to poor loop closure detection.
The key insight here is that direct VO does not care about the \emph{repeatability} of the selected (or tracked) pixels. 
Thus, direct VO systems have in the past been extended to SLAM either by using only keyframe proximity for loop closure detection \cite{Kerl2013} or by computing features for loop closure detection independently from frontend tracking and constraint computation \cite{Engel2014}.
Direct image alignment is then used to estimate relative pose constraints \cite{Kerl2013, Engel2014}, which requires images of keyframes to be kept available.
We propose instead to gear point selection towards repeatable features and use geometric techniques to estimate constraints.
In summary, our contributions are:
\begin{itemize}
\item We adapt DSO's point selection strategy to favor repeatable corner features, while retaining its robustness against feature-poor environments. The selected corner features are then used for loop closure detection with conventional BoW.
\item We utilize the depth estimates of matched feature points to compute $\Sim(3)$ pose constraints with a combination of pose-only bundle adjustment and point cloud alignment, and --- in parallel to the odometry frontend --- fuse them with a co-visibility graph of relative poses extracted from DSO's sliding window optimization.
\item We demonstrate on publicly available real-world datasets that the point selection retains the tracking frontend's accuracy and robustness, and the pose graph optimization significantly reduces the odometry's drift and results in overall performance comparable state-of-the-art feature-based methods, even without global bundle adjustment.
\item We make our implementation publicly available\footnote{\url{https://vision.in.tum.de/research/vslam/ldso}}.
\end{itemize}
Fig.~\ref{fig:1}~illustrates how LDSO corrects accumulated drift after closing a loop in the TUM-Mono dataset.

%%%%%%%%%%%%%%%%%%%%%%%%%%%%%%%%%%%%%%%%%%%%%%%%%%%%%%%%%%%%%%%%%%%%%%%%%%%%%%%%
\section{RELATED WORK}

Many mature feature-based monocular SLAM systems have been presented in recent years, often inspired by the seminal PTAM \cite{Klein2007}, where splitting the system into a camera tracking frontend and an optimization-based mapping backend was originally proposed, and later ScaViSLAM \cite{Strasdat2011}, which suggested to mix local bundle adjustment with $\Sim(3)$ pose-graph optimization.
One of the more influential such systems has been ORB-SLAM \cite{Mur-Artal2015}. 
It features multiple levels of map-optimization, starting from local bundle-adjustment after keyframe insertion, global pose-graph optimization after loop closures detected with BoW, and finally (expensive) global bundle adjustment. 
Since unlike LDSO it uses traditional feature matching to localize images against the current map, much emphasis is on map-maintenance by removing unneeded keyframes and unused features. Note that while loop-closure detection and pose graph optimization are similar in LDSO, we only need to compute feature descriptors for keyframes.

Visual odometry systems have been extended to SLAM systems in different ways. It is interesting to note that many systems propose to integrate inertial sensors \cite{Leutenegger2015, Qin2017, Lynen2015, Schneider2018} and/or use a stereo setup \cite{Leutenegger2015, Schneider2018, mur2017orb, Wang2017}, since this can increase robustness in challenging environments and make additional degrees of freedom observable (global scale, roll and pitch).
Lynen et al. \cite{Lynen2015} proposes to directly include 2D-3D matches from an existing map as Kalman Filter updates in a local MSCKF-style odometry estimator.
Okvis \cite{Leutenegger2015} is a feature-based visual-inertial keyframe odometry that maintains a local map of feature points in a constant-size marginalization window (similar to DSO) that does not reuse points once it is out of the local window.
Later, a similar visual-inertial weighted least-squares optimization strategy was adapted in maplab \cite{Schneider2018} for batch map optimization without marginalization, but as a camera tracking frontend, a Kalman Filter based estimator minimizing the photometric error between tracked image patches is used. 
Feature-based localization against an existing map is incorporated in the frontend as pose updates.
VINS-Mono \cite{Qin2017} in turn is a feature-based monocular visual-inertial SLAM system that employs a marginalizing odometry front-end very similar to okvis. 
It includes feature observations from an existing map in this sliding window optimization. Similar to ORB-SLAM and our work, loop closure and global map refinement are based on BoW and pose graph optimization, but with help of the inertial sensors, it suffices to use non-rotation-invariant BRIEF descriptors and do pose graph optimization in 4 degrees-of-freedom. 
While in ORB-SLAM the feature extraction step costs almost half of the running time, the frontend tracking in VINS-Mono is based on KLT features and thus is capable of running in real-time on low-cost embedded systems. 
This however means, that for loop closure detection additional feature points and descriptors have to be computed for keyframes.

As a direct monocular SLAM system and predecessor of DSO, LSD-SLAM \cite{Engel2014} employs FAB-MAP \cite{Cummins2008} --- an appearance-only loop detection algorithm --- to propose candidates for large loop closures.
However, FAB-MAP needs to extract its own features and cannot re-use any information from the VO frontend, and the constraint computation in turn does not re-use the feature matches, but relies on direct image alignment using the semi-dense depth maps of candidate frames in both directions and a statistical test to verify the validity of the loop closure, which also means that images of all previous keyframes need to be kept available.

%%%%%%%%%%%%%%%%%%%%%%%%%%%%%%%%%%%%%%%%%%%%%%%%%%%%%%%%%%%%%%%%%%%%%%%%%%%%%%%%
\section{LOOP CLOSING IN DSO}
\subsection{Framework}
Before delving into the details of how our loop closing thread works, we first 
briefly introduce the general framework and formulation of DSO. DSO is a 
keyframe-based sliding window approach, where 5-7 keyframes are maintained and 
their parameters are jointly optimized in the current window. Let $\mathcal{W} 
= \{\mathbf{T}_1, \ldots, \mathbf{T}_m, \mathbf{p}_1, \ldots, \mathbf{p}_n\}$ 
be 
the $m$ $\SE(3)$ keyframe poses and $n$ points (inverse depth parameterization) in the sliding window, the photometric error to be 
minimized is defined as~\cite{Engel2018}:
\begin{equation}\label{eq:photometric-cost}
\begin{aligned}
\min & \sum\limits_{{\mathbf{T}_i},{\mathbf{T}_j},{\mathbf{p}_k} \in 
\mathcal{W}} 
{{E_{i,j,k}}}, \quad \text{where} \\
{{E_{i,j,k}}} &= {\sum\limits_{\mathbf{p} \in {\mathcal{N}_{\mathbf{p}_k}}} 
{{w_\mathbf{p}}\left\| {\left( {{I_j}[\mathbf{p}'] - {b_j}} \right) - \frac{t_j 
e^{a_j}}{t_i e^{a_i}}\left( {{I_i}[\mathbf{p}] - {b_i}} \right)} 
\right\|_\gamma}},
\end{aligned}
\end{equation}
where $\mathcal{N}_{\mathbf{p}_k}$ is the neighborhood pattern of $\mathbf{p}_k$, $a,b$ 
are the affine light transform parameters, $t$ is the exposure time, $I$ denotes an 
image and $w_{\mathbf{p}}$ is a heuristic weighting factor. $\mathbf{p}'$ is 
the reprojected pixel of $\mathbf{p}$ on $I_j$ calculated by
\begin{equation}\label{eq:warping}
\begin{aligned}
\mathbf{p}'=\Pi(\mathbf{R}\Pi^{-1}(\mathbf{p}, d_{\mathbf{p}_k})+\mathbf{t}),
\end{aligned}
\end{equation}
with $\Pi:\mathbb{R}^{3} \to \Omega$ the projection and $\Pi^{-1}:\Omega \times 
\mathbb{R} \to \mathbb{R}^{3}$ the back-projection function, $\mathbf{R}$ and 
$\mathbf{t}$ the 
relative rigid body motion between the two frames calculated from 
$\mathbf{T}_i$ and 
$\mathbf{T}_j$, $d$ the inverse depth of the point.

As a new frame arrives, DSO estimates its initial pose using direct image alignment 
by projecting all the active 3D points in the current window into this frame. 
If required, this frame thereafter will be added into the local windowed bundle 
adjustment. The sliding window naturally forms a co-visibility graph like in 
ORB-SLAM, but the co-visible information is never used outside the local 
window, as old or redundant keyframes and points are marginalized out. Although the windowed optimization becomes computationally 
light-weight and more accurate using the marginalization prior, the estimation will 
inevitably drift.

A global optimization pipeline is needed in order to close long-term loops for 
DSO. Ideally global bundle adjustment using photometric error should be used, 
which nicely would match the original formulation of DSO. However, in that case all 
the images would need to be saved, since the photometric error is computed on images. 
Moreover, nowadays it is still impractical to perform global photometric bundle adjustment 
for the amount of points selected by DSO. To avoid these problems we turn to 
the idea of using pose graph optimization, which leaves us several other 
challenges: (i) How to combine the result of global pose graph optimization 
with that of the windowed optimization? One step further, how to set up the 
pose graph constraints using the information in the sliding window, considering 
that pose graph optimization minimize $\Sim(3)$ geometry error between keyframes 
while in the sliding window we minimize the photometric error? (ii) How to 
propose loop candidates? While the mainstream of loop detection is based on 
image descriptors, shall we simply add another thread to 
perform those feature related computations? (iii) Once loop candidates are proposed we need 
to compute their relative $\Sim(3)$ transformation. In 
a direct image alignment approach, we need to set an initial guess on the 
relative pose to start the Gauss-Newton or the Levenberg-Marquardt iterations, 
which is challenging in this case as the relative motion may be far away from identity.

\begin{figure}[!t]
	\centering
	% figure 2. framework
	\includegraphics[width=0.4\textwidth]{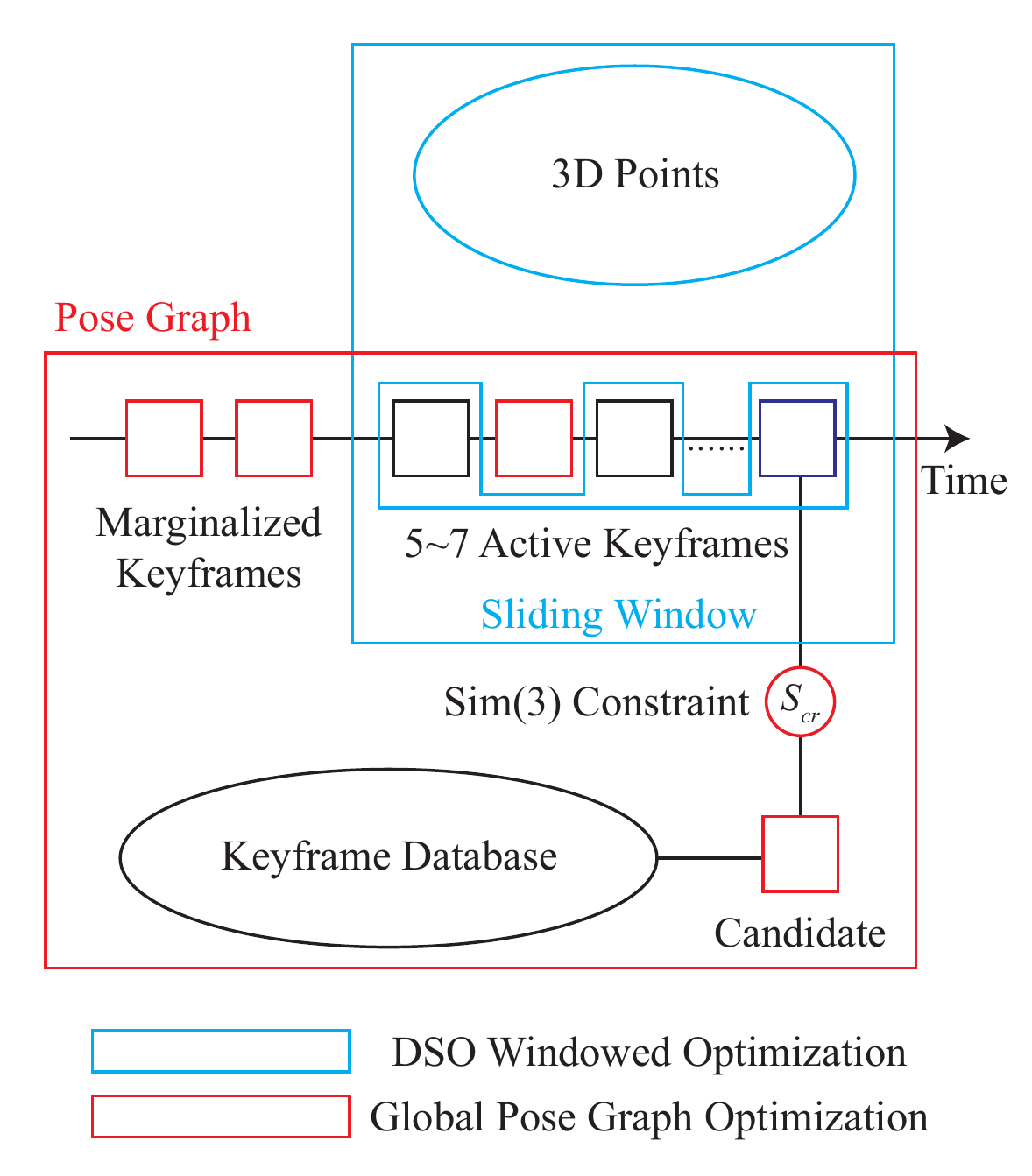}
	\vspace{-1em}
	\caption{Framework of the proposed system. }
	\label{fig:2}
	\vspace{-1em}
\end{figure}

% figure, pixel selection in DSO and DSO_LP 
\begin{figure*}[!t]
	\centering
	\includegraphics[width=0.98\textwidth]{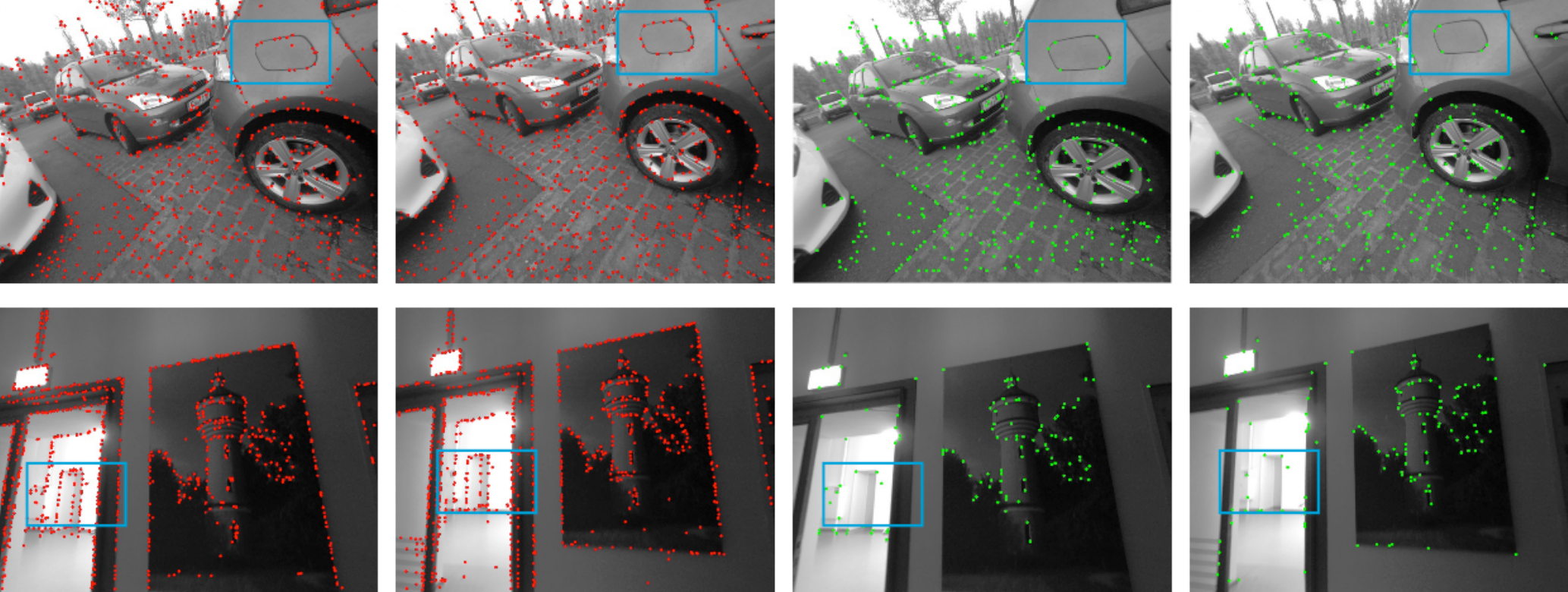}
	\caption{Pixel selection in DSO and LDSO. The left part is pixels picked by 
		DSO and the right part shows the corners of LDSO (we don't show the 
		non-corners to make the image look clear). And the top part is a well 
		textured environment and the bottom part is a weak textured one. Note that 
		if the number of corners is less than the threshold, we will also pick 
		extra pixels just like DSO to make the system robust against feature-less 
		situations. The points in the blue box shows the difference of point repeatability. }
	\label{fig:pixel-select}
\end{figure*}

Taking these challenges into account, we design our loop closing module as 
depicted in Fig.~\ref{fig:2}. Alongside the DSO window, we add a global pose 
graph to maintain the connectivity between keyframes. 
DSO's sliding window naturally forms a co-visibility graph where we can 
take the relative 3D pose transformations between the keyframes as the pairwise 
pose measurements. For loop detection and validation, we rely on BoW and 
propose a novel way to combine ORB features with the original sampled points of 
DSO. In this way, if a loop candidate is proposed and validated, its $\Sim(3)$ 
constraint with respect to the current keyframe is calculated and added to 
the global pose graph, which is thereafter optimized to obtain a more 
accurate long-term camera pose estimation.

\subsection{Point Selection with Repeatable Features}
It is worth noting that even in direct methods like LSD-SLAM or DSO, 
point selection is still needed; One difference from indirect methods is that 
the repeatability of those points is not required by direct methods. 
DSO uses a dynamic grid search to pick enough 
pixels even in weakly textured environments. We modify this strategy to make it 
more sensitive to \emph{corners}. More specifically, we still pick a given 
number of pixels (by default 2000 in DSO), in which part of them are corners 
(detected by using the easy-to-compute Shi-Tomasi score~\cite{Shi1994}), while 
the others are still selected using the method proposed for DSO. Keeping the 
number of corners small,  we compute their ORB 
descriptors~\cite{Rublee2011} and pack them into BoW. The VO frontend uses both 
the corners and the non-corners for camera tracking, keeping therefore the 
extra overhead for feature extraction of the loop closing thread to a minimum.

Fig.~\ref{fig:pixel-select} shows the pixel selection in the original DSO and 
the LDSO. It can be seen that the pixels picked by DSO have little 
repeatability and therefore it is hard to seek image matchings using those 
points for loop closure. 
In LDSO we use both corners and other pixels with high gradients, where 
the corners are used both for building BoW models and for tracking, while the 
non-corners are only used for tracking. In this way we can on one hand track 
the weak texture area, on the other hand also match features between keyframes 
if needed. 

\subsection{Loop Candidates Proposal and Checking}
As we compute ORB descriptors for each keyframe, a BoW database is built
using DBoW3~\cite{Galvez-Lopez2012}. 
Loop candidates are proposed for the current keyframe by querying the database 
and we only pick those that are outside the current window (i.e., marginalized 
keyframes). For each candidate we try to match its ORB features to those of the 
current keyframe, and then perform RANSAC PnP to compute an initial guess of the 
$\SE(3)$ transformation. Afterwards we optimize a $\Sim(3)$ transformation using  
Gauss-Newton method by minimizing the 3D and 2D geometric constraints. Let $\mathcal{P}=\{\mathbf{p}_i\}$ 
be the reconstructed features in the loop candidate and $d_{\mathbf{p}_i}$ their 
inverse depth, $\mathcal{Q}=\{\mathbf{q}_i\}$ be the matched features in the 
current keyframe, 
$\mathbf{D}$ be the sparse inverse depth map of the current keyframe which is computed 
by projecting the active map points in the current window into the current keyframe. For 
some features in $\mathcal{Q}$ we can find their depth in $\mathbf{D}$, for the others we 
only have their 2D pixel positions. Let $\mathcal{Q}_1 \subseteq \mathcal{Q}$ be 
those without depth and $\mathcal{Q}_2=\mathcal{Q}\backslash \mathcal{Q}_1$ be those with depth, 
then the $\Sim(3)$ transformation from the loop candidate 
(reference) to the current keyframe $\mathbf{S}_{cr}$ can be estimated by minimizing the 
following cost function:
\begin{equation}
\begin{split}
{\mathbf{E}_{loop}} = & \sum\limits_{{\mathbf{q}_i} \in {\mathcal{Q}_1}} {w_1}{{\left\| 
{{\mathbf{S}_{cr}}{\Pi^{-1}(\mathbf{p}_i, d_{\mathbf{p}_i})} - \Pi^{-1}(\mathbf{q}_i, 
d_{\mathbf{q}_i})} \right\|}_2} + \\ 
& \sum\limits_{{\mathbf{q}_j} \in {\mathcal{Q}_2}} {{w_2}{{\left\| {\Pi 
({\mathbf{S}_{cr}}\Pi^{-1}(\mathbf{p}_j, d_{\mathbf{p}_j})) - {\mathbf{q}_j}} \right\|}_2}} ,
\end{split}
\end{equation}
where $\Pi(\cdot)$ and $\Pi^{-1}(\cdot)$ are the projection and back-projection functions 
as defined before, $w_1$ and $w_2$ are weights to balance the 
different measurement units. In practice the scale can only be estimated by the 3D part, 
but without the 2D reprojection error, the rotation and translation estimate will be inaccurate
if the estimated depth values are noisy.

\subsection{Sliding Window and $\Sim(3)$ Pose Graph}
In this section we explain how to fuse the estimations of the sliding window 
and the global pose graph. Let $\mathbf{x} = [\mathbf{x}_p^\top, \mathbf{x}_d^\top]^\top$ with 
$\mathbf{x}_p$ the poses of the keyframes in 
the current window parameterized using twist coordinates, $\mathbf{x}_d$ the 
points parameterized by their inverse depth $d$, then the windowed optimization 
problem using Levenburg-Marquardt (L-M) iterations is: 
\begin{equation}
	\mathbf{H} \delta \mathbf{x} =  - \mathbf{b},
\end{equation}
where $\mathbf{H}$ is the Hessian matrix approximated as $\mathbf{J}^\top 
\mathbf{WJ} + \lambda \mathbf{I}$ in L-M iterations, $\delta 
\mathbf{x}$ is the optimal increment, $\mathbf{W}$ is a weight matrix, 
$\mathbf{b}=\mathbf{J}^\top\mathbf{W}\mathbf{r}$ with Jacobian $\mathbf{J}$ and residual $\mathbf{r}$. 
It can also be written 
in a block-matrix way
\begin{equation}
{\begin{bmatrix}
{{\mathbf{H}_{pp}}}&{{\mathbf{H}_{pd}}}\\
{{\mathbf{H}_{dp}}}&{{\mathbf{H}_{dd}}}
\end{bmatrix}}{\begin{bmatrix}
{\delta {\mathbf{x}_p}}\\
{\delta {\mathbf{x}_d}}
\end{bmatrix}} =  - {\begin{bmatrix}
{{\mathbf{b}_p}}\\
{{\mathbf{b}_d}}
\end{bmatrix}}.
\end{equation}
It is well known that $\mathbf{H}$ has an arrow-like sparse pattern (in DSO's 
formulation $\mathbf{H}_{dd}$ being a diagonal matrix) where we can exploit the 
sparsity in the bottom right part to perform sparse bundle adjustment 
\cite{Sibley2008, Triggs2000}. 

\begin{figure*}[!t]
	\centering
	\includegraphics[width=0.98\textwidth]{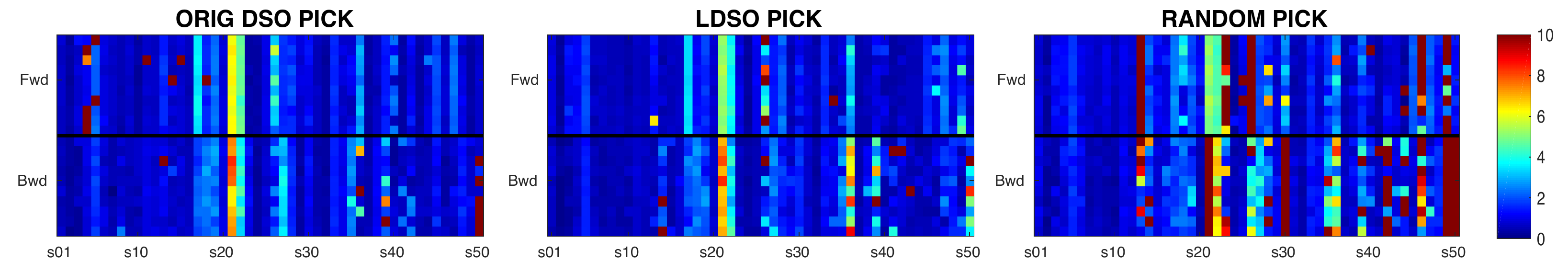}
	\vspace{-1em}
	\caption{The alignment errors $e_{\text{align}}$ using different 
		points picking 
		strategies. Each small square block is the color-coded alignment error 
		for one run as defined in~\cite{Engel2016a} and each column 
		corresponds to each sequence of the dataset. We run our method 10 times 
		forward and 10 times backward on each sequence to account for the 
		nondeterministic behavior.}
	\label{fig:block-pick}
	\vspace{-1em}
\end{figure*}

The marginalization strategy in DSO keeps the sparsity pattern in 
$\mathbf{H}_{dd}$ and also keeps a motion prior expressed as a quadratic 
function on $\mathbf{x}$ {(for details please refer to Eq. (19) 
in~\cite{Engel2018})}. This prior can be also regarded as a hyper edge in the 
pose graph which constraints all the keyframes inside. However, traditional 
pose graph optimization takes only pair-wise observations between two keyframes 
like $\mathbf{T}_{ij}$ and compute the measurement error:
\begin{equation}
	\mathbf{e}_{ij} = \mathbf{T}_{ij} \hat{\mathbf{T}}_j^{-1} 
	\hat{\mathbf{T}}_i,
\end{equation}
where $\hat{(\cdot)}$ is the estimated value of a variable.

Since our loop closing approach computes relative pose constraints between the loop 
candidate and the current frame, we also approximate the constraints inside the marginalization window 
with pairwise relative pose observations. Specifically, we compute those 
observations from the frontend's current global pose estimates.

It is also important to note that, since we do not want to disturb the local 
windowed optimization (it contains absolute pose information), in pose graph 
optimization we will fix the current frame's pose estimation. Therefore, the 
pose graph optimization will tend to modify the global poses of the old part of 
the trajectory. Besides, the global poses of the keyframes in the current 
window are not updated after the pose graph optimization, to further make sure 
that the local windowed bundle adjustment is not influenced by the global 
optimization. Our implementation is based on g2o, a graph optimization library 
proposed in~\cite{Kummerle2011}.

%%%%%%%%%%%%%%%%%%%%%%%%%%%%%%%%%%%%%%%%%%%%%%%%%%%%%%%%%%%%%%%%%%%%%%%%%%%%%%%%
\section{EVALUATION}

We evaluate our method on three popular public datasets: TUM-Mono~\cite{Engel2016a}, 
EuRoC MAV~\cite{Burri25012016} and KITTI Odometry~\cite{Geiger2012}, all in a 
monocular setting. 

\subsection{The TUM-Mono Dataset}
The TUM-Mono Dataset is a monocular dataset that consists of 50 indoor and 
outdoor sequences. It provides photometric camera calibration, but no full 
ground-truth camera trajectories. The camera always returns to the starting 
point in all sequences, making this dataset very suitable for evaluating 
accumulated drifts of VO systems. For this reason we disable the loop closure 
functionality of our method on this dataset, to first evaluate the VO accuracy 
of our method with the modified point selection strategy.

We evaluate three different point selection strategies: (1) random point 
selection; (2) the original method of DSO and (3) our method. 
For each strategy we run 10 times forward and 10 times backward on each 
sequence to account for the nondeterministic behavior. We compute the 
accumulated translational, rotational and scale drifts $e_t$, $e_r$, $e_s$ in 
the keyframe trajectories using method described in \cite{Engel2016a}. 

\begin{figure}[!t]
	\centering
	\includegraphics[width=0.5\textwidth]{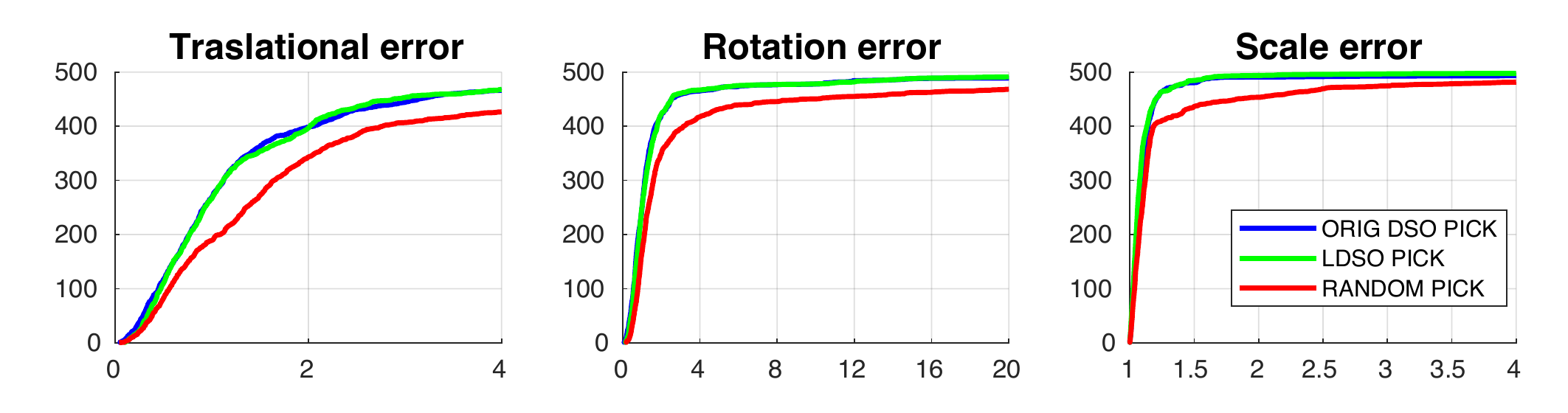}
	\caption{Accumulated translational, rotational and scale 
		drifts using 
		different points picking strategies. X-axis is the error threshold, and 
		Y-axis is the number of runs whose errors are below the threshold. }
	\label{fig:RTS-pick}
\end{figure}

Fig.~\ref{fig:block-pick} shows the color-coded alignment error in all the 
sequences. Fig.\ref{fig:RTS-pick} shows the cumulative error plot, which depicts 
the number of runs whose errors are below the corresponding x-values (thus closer to left-top is better). 
In both figures we see that our integration of corner features into DSO does 
not reduce the VO accuracy of the original system. Another interesting point 
is, although random picking makes the tracking fail more frequently, it seems 
it does not increase the errors on those successfully tracked sequences like 
s01 to s10.

\begin{figure}
	\centering	
	\includegraphics[width=0.48\textwidth]{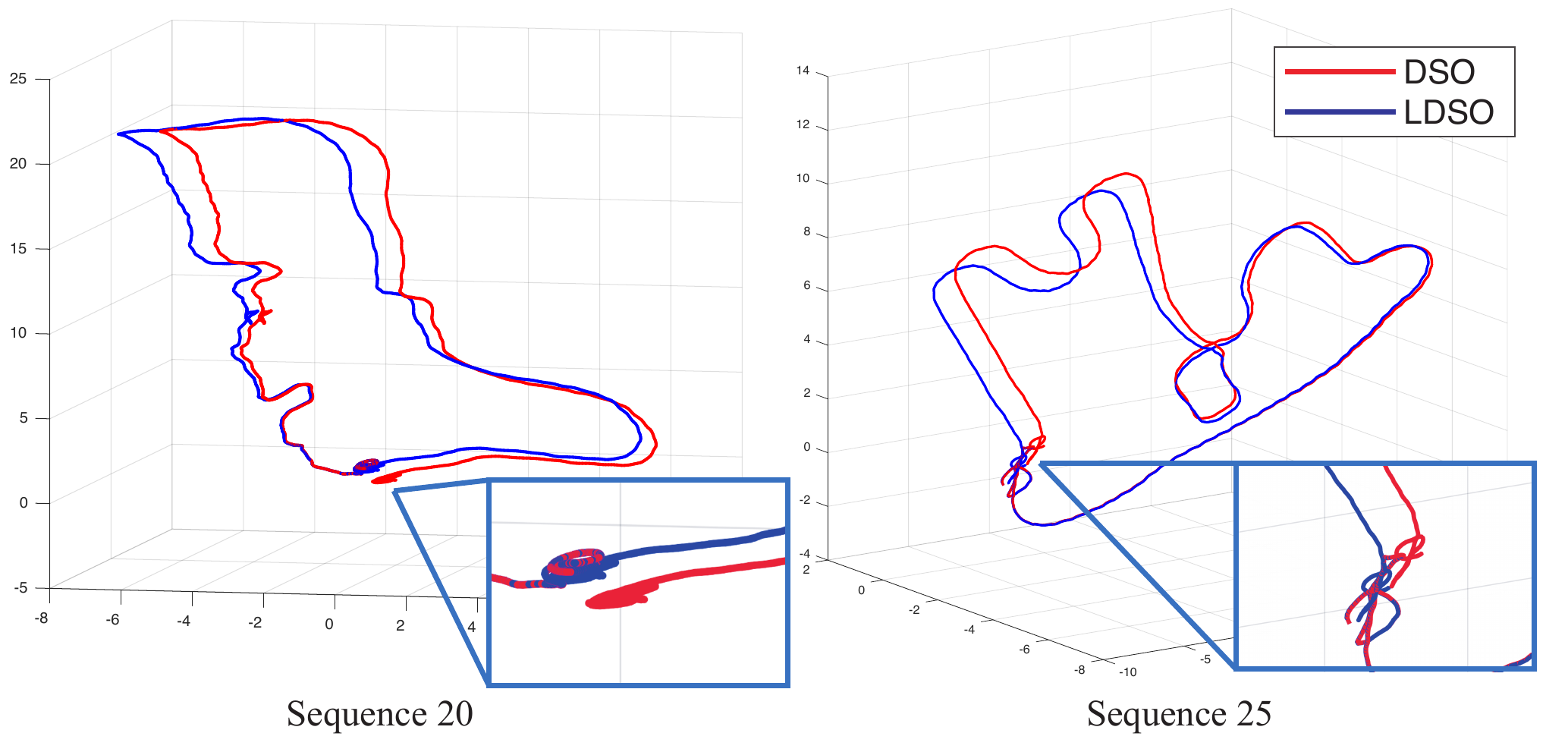}
	\vspace{-1em}
	\caption{Trajectories in TUM-Mono dataset. The red line is the estimated 
		trajectory by original DSO where we can see obvious drift. The blue 
		line is 
		the loop-closed trajectory. }
	\label{fig:traj}
\end{figure}

\begin{figure}
	\centering	
	\includegraphics[width=0.48\textwidth]{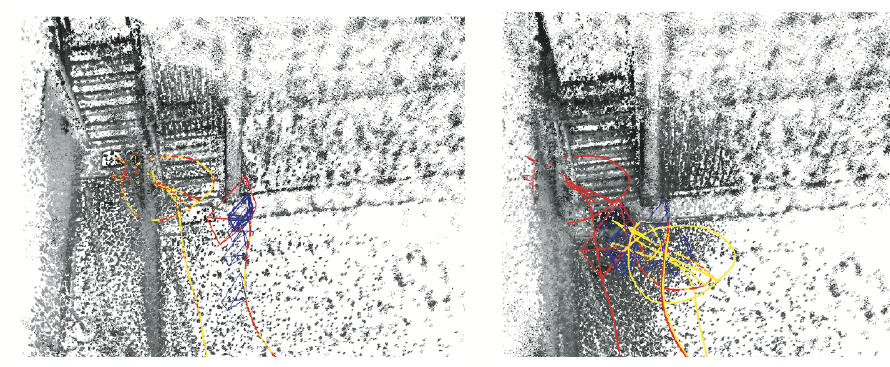}
	\caption{Map before and after loop closure in LDSO (sequence 33). }
	\label{fig:map-compare}
\end{figure}

To show some qualitative results we also run LDSO with loop closing on TUM-Mono 
and get some $\Sim(3)$ closed trajectories shown in Fig.~\ref{fig:traj}. An 
example of the reconstructed map is shown in Fig.~\ref{fig:map-compare}.

\subsection{The EuRoC MAV Dataset}
\begin{figure}[!t]
	\centering
	\includegraphics[width=0.48\textwidth]{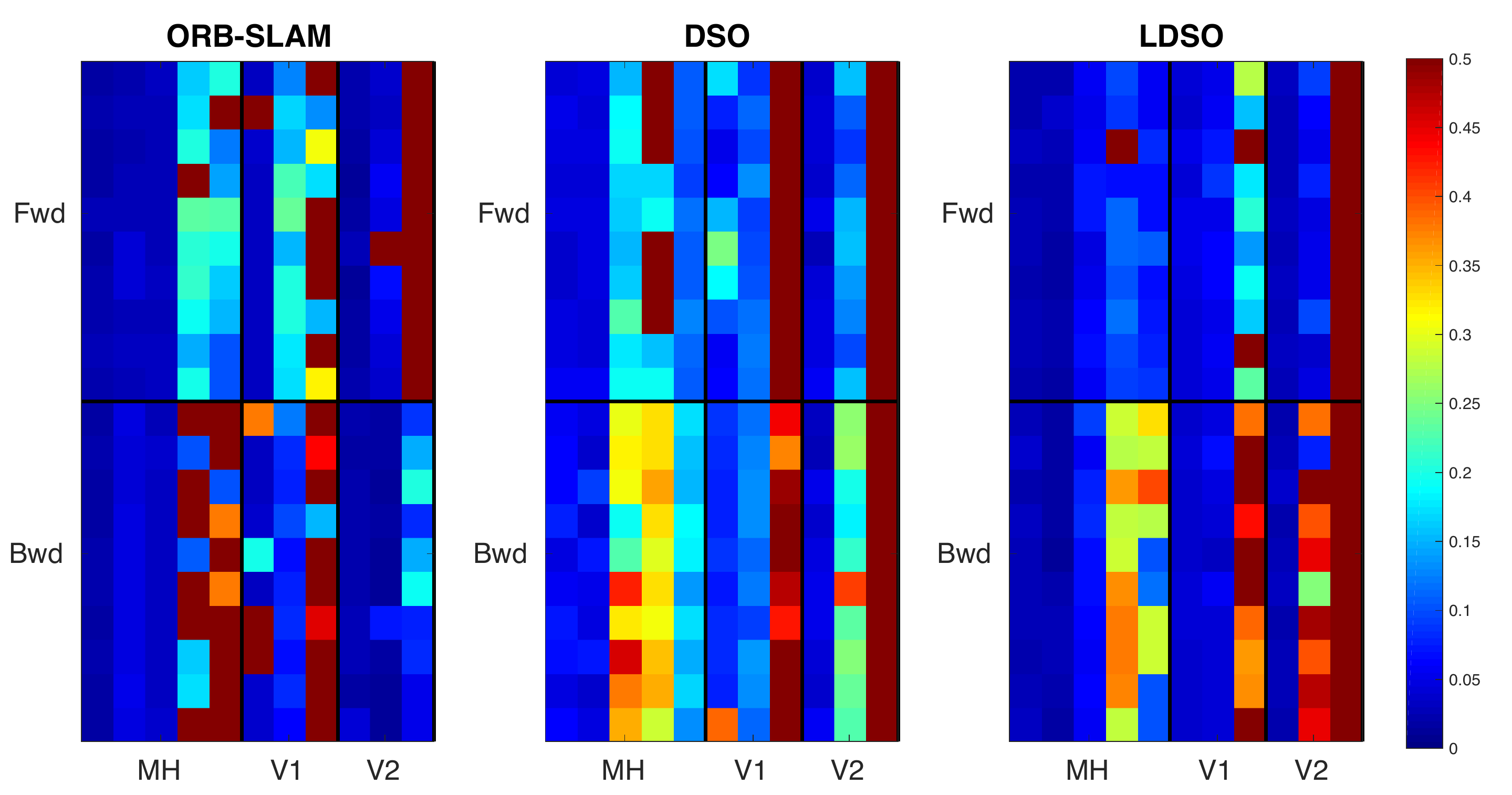}
	\vspace{-1em}
	\caption{Full trajectory RMSE of all sequences ($\Sim(3)$ aligned to the 
		ground truth). X-axis is the sequence name, which varies from MH\_01 to 
		MH\_05 and then V1\_01 to V2\_03.}
	\label{fig:EUROC-block}
	\vspace{-1em}
\end{figure}

\begin{figure}
	\centering
	\includegraphics[width=0.40\textwidth]{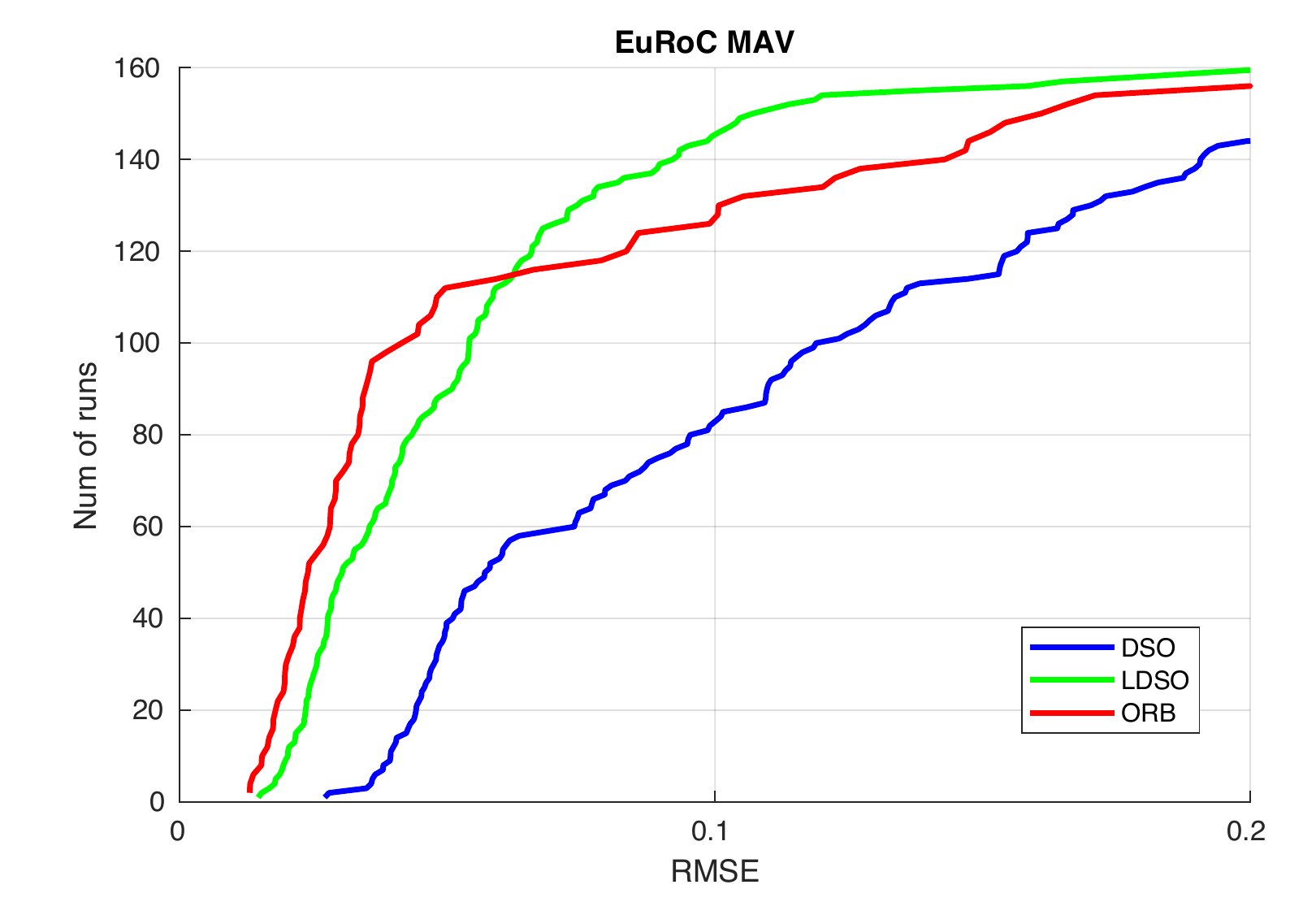}
	\vspace{-1em}
	\caption{RMSE on EuRoC MAV. The Y-axis shows the number of runs with errors 
		below the corresponding values on the X-axis. All errors are calculated 
		after $\Sim(3)$ alignment of the camera trajectories to the ground truth.}
	\label{fig:EUROC-RMSE}
	\vspace{-1em}
\end{figure}

The EuRoC MAV Dataset provides 11 sequences with stereo images, synchronized 
IMU readings and ground-truth camera trajectories. We compare LDSO with DSO and 
ORB on this dataset by evaluating their root-mean-square error (RMSE) using 
their monocular settings. Same as before on each sequence we run 10 times 
forward and 10 times backward for each method and the results are shown in 
Fig.~\ref{fig:EUROC-block} and Fig.~\ref{fig:EUROC-RMSE}. Generally speaking, 
ORB-SLAM2 performs quite well on this dataset and it only fails consistently on 
sequence V2-03 when running forward. DSO and LDSO both fail on sequence V2-03, 
but on most of the others sequences LDSO significantly improves the camera 
tracking accuracy. The overall improvement after having loop closure can also 
be found in Fig.~\ref{fig:EUROC-RMSE}. From the plot we can see that
ORB-SLAM2 is more accurate, whereas LDSO is more robust on this dataset.

\subsection{The KITTI Odometry Dataset}
On the KITTI Odometry Dataset, as shown previously in the thorough evaluation 
in~\cite{Wang2017}, monocular VO systems like DSO and ORB-SLAM (the VO 
component only) suffer severe accumulated drift which makes them not usable 
for such large-scale scenarios. While the natural way to resolve this problem 
is to integrate other sensors like IMU or to use stereo cameras which has been 
proved quite successful~\cite{Wang2017, mur2017orb}, here we want to see the 
potential of our monocular method after the integration of the loop closure 
functionality.  

\begin{figure*}
	\centering
	\includegraphics[width=0.94\textwidth]{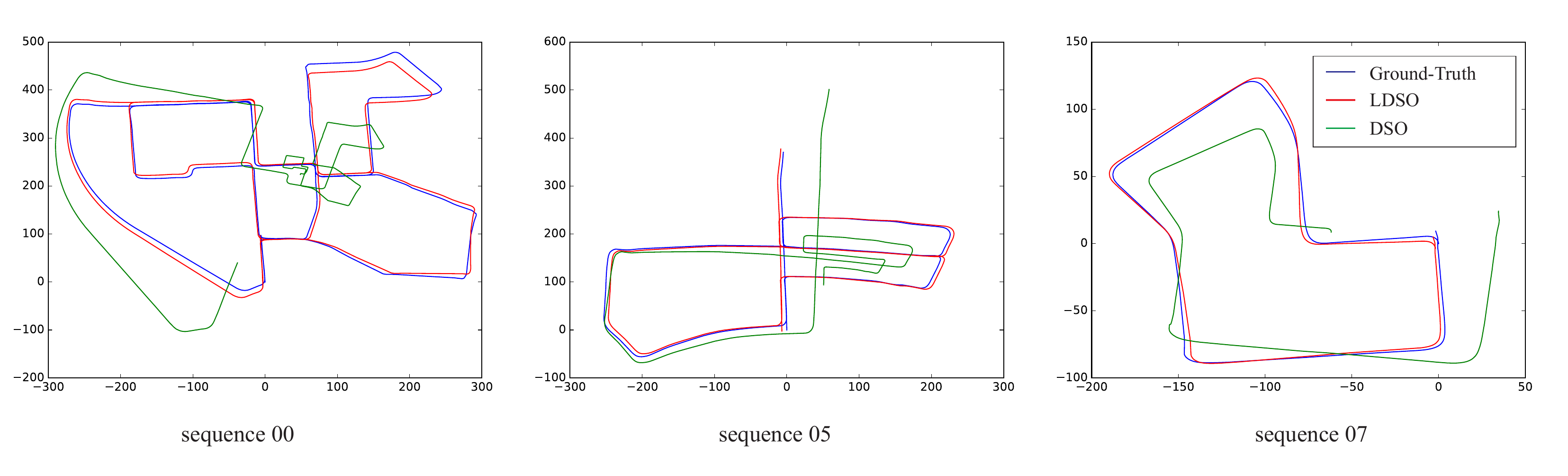}
	\vspace{-1.5em}
	\caption{$\Sim(3)$ aligned trajectories of Kitti sequence 00, 05 and 07, which contains 
	closed loops.}
	\vspace{-1.5em}
	\label{fig:Kitti-traj}
\end{figure*}

We compare LDSO with DSO and ORB-SLAM2 and show the Absolute Trajectory 
Errors (ATEs) on all the sequences of the training set in 
Table~\ref{table:ATE-Kitti}. The ATEs are computed by performing $\Sim(3)$ 
alignment to the ground-truth. Not surprisingly on sequences with loops 
(seq. 00, 05, 07), LDSO improves the performance of DSO a lot. Besides,
our method achieves comparable accuracy to ORB-SLAM2, which has a global bundle 
adjustment in the loop closing thread and we only use pose graph optimization. 
Some qualitative results on the estimated camera trajectories can be found in 
Fig.~\ref{fig:Kitti-traj}.
% accumulated trans error

% trajectory figure

% map figure

\begin{table}[!htb]
	\centering
	\caption{ATE error (m) on all Kitti training sequences.}
\begin{tabular}{c|ccc}
	\hline \hline
	Sequence & Mono DSO & LDSO & ORB-SLAM2 \\ 
	\hline 
	00 & 126.7 & 9.322 & \textbf{8.27} \\ 
	01 & 165.03 & \textbf{11.68} & x \\ 
	02 & 138.7 & 31.98 & \textbf{26.86} \\ 
	03 & 4.77 & 2.85 & \textbf{1.21} \\ 
	04 & 1.08 & 1.22 & \textbf{0.77} \\ 
	05 & 49.85 & \textbf{5.1} & 7.91 \\  
	06 & 113.57 & 13.55 & \textbf{12.54} \\ 
	07 & 27.99 & \textbf{2.96} & 3.44 \\ 
	08 & 120.17 & 129.02 & \textbf{46.8}1 \\ 
	09 & 74.29 & \textbf{21.64} & 76.54 \\
	10 & 16.32 & 17.36 & \textbf{6.61} \\\hline\hline
\end{tabular} 
\vspace{-5em}
	\label{table:ATE-Kitti}
\end{table}

\subsection{Runtime Evaluation}
Finally we present a short runtime analysis about the point selection step. Note that 
loop closure only occurs very occasionally and the pose graph is running in a single 
thread, thus they do not affect much the computation time of the main thread. What we 
change in the main thread is adding an extra feature extraction and descriptor 
computation step. But unlike the feature-based approaches, they are not performed for 
every frame but only for keyframes. 
Table \ref{table:computation-time} shows the average computation time  of the point 
selection step using different picking strategies. 
The point selection in LDSO takes slightly more time than that in DSO due to the feature 
and descriptor extraction. It is worth noting that the values are calculated over keyframes, 
thus the runtime impact will be further moderated when averaging over all frames.
The program is tested on a laptop with Ubuntu 18.04 and Intel i7-4770HQ CPU and 16GB 
RAM. 

\begin{table}[!htp]
	\centering
	\vspace{1em}
	\caption{Runtime of Different Point Selection Strategies}
	\begin{tabular}{ccc}
		\hline\hline
		 LDSO Pick & DSO Pick & Random Pick \\
		 0.0218s & 0.0126s & 0.0027s \\
		 \hline\hline
	\end{tabular}
	\label{table:computation-time}
	\vspace{-1em}
\end{table}

%%%%%%%%%%%%%%%%%%%%%%%%%%%%%%%%%%%%%%%%%%%%%%%%%%%%%%%%%%%%%%%%%%%%%%%%%%%%%%%%
\section{CONCLUSION}
In this paper we propose an approach to integrate loop closure and global map optimization into the fully direct VO 
system DSO. 
DSO's original point selection is adapted to include repeatable features. For those we compute ORB descriptors and build BoW models for loop closure detection. 
We demonstrate that the point selection retains the original robustness and accuracy of the odometry frontend, while enabling the backend to effectively reduce global drift in rotation, translation and scale.
We believe the proposed approach can be extended to future improvements of VO or SLAM. 
For example, a photometric bundle adjustment layer might increase the global map accuracy.
In order to ensure long-term operation, map maintenance strategies such as keyframe culling and removal of redundant 3D points may be employed.
Combining the information from 3D points of neighboring keyframes after loop closure may help to further increase the accuracy of the reconstructed geometry.

\balance

%%%%%%%%%%%%%%%%%%%%%%%%%%%%%%%%%%%%%%%%%%%%%%%%%%%%%%%%%%%%%%%%%%%%%%%%%%%%%%%%%
%\section*{APPENDIX}
%Here is appendix.
%
%\section*{ACKNOWLEDGMENT}
%Here is acknowledgment.
%
%%%%%%%%%%%%%%%%%%%%%%%%%%%%%%%%%%%%%%%%%%%%%%%%%%%%%%%%%%%%%%%%%%%%%%%%%%%%%%%%%
\bibliographystyle{ieeetr}
\bibliography{./ref}

\end{document}